\documentclass[sigconf]{acmart}
\AtBeginDocument{%
  }

\copyrightyear{2026}
\acmYear{2026}
\setcopyright{cc}
\setcctype{by}
\acmConference[KDD '26]{Proceedings of the 32nd ACM SIGKDD Conference on Knowledge Discovery and Data Mining V.2}{August 09--13, 2026}{Jeju Island, Republic of Korea}
\acmBooktitle{Proceedings of the 32nd ACM SIGKDD Conference on Knowledge Discovery and Data Mining V.2 (KDD '26), August 09--13, 2026, Jeju Island, Republic of Korea}
\acmDOI{10.1145/3770855.3818984}
\acmISBN{979-8-4007-2259-2/2026/08}

\usepackage{algorithm, algpseudocode, makecell, enumitem, listings, multicol, multirow, tikz, fontawesome5}
\usepackage[dvipsnames]{xcolor}
\usepackage[most]{tcolorbox}
\usepackage{float}

\newcommand{\inc}[1]{\scriptsize\textcolor{ForestGreen}{(+#1)}}
\newcommand{\dec}[1]{\scriptsize\textcolor{red}{(#1)}}
\newcommand{\circled}[1]{\tikz[baseline=(char.base)]{\node[shape=circle, draw, fill=black, inner sep=0.8pt] (char) {\color{white}\small #1};}}

\algrenewcommand\algorithmicrequire{\textbf{Input:}}
\algrenewcommand\algorithmicensure{\textbf{Output:}}

\tcbuselibrary{listings, skins, breakable}
\definecolor{codebg}{RGB}{245,245,245}
\definecolor{calcbg}{RGB}{15,15,112}
\lstdefinestyle{promptstyle}{
  basicstyle=\ttfamily\scriptsize,
  columns=fullflexible,
  breaklines=true,
  breakatwhitespace=true,
  breakindent=0pt,
  breakautoindent=false,
  keepspaces=true,
  showstringspaces=false,
  xleftmargin=0pt,
  framexleftmargin=0pt,
  aboveskip=0pt,
  belowskip=0pt
}

\newtcblisting{prompt}[2][]{
  breakable,
  enhanced,
  colback=black!1,
  colframe=black!40,
  boxrule=0.6pt,
  arc=2pt,
  left=5pt,
  right=5pt,
  top=5pt,
  bottom=5pt,
  fonttitle=\bfseries,
  title={#2},
  listing only,
  listing options={
    style=promptstyle
  },
  #1
}

\begin{document}
\raggedbottom

\title{Redefining Maritime Anomaly Detection via Equation-Grounded Synthetic Anomalies}

\author{Youngseok Hwang}
\authornote{Both authors contributed equally to this research.}
\email{yshwang35@snu.ac.kr}
\affiliation{%
  \institution{Seoul National University}
  \city{Seoul}
  \country{Republic of Korea}
}

\author{Sungho Bae}
\authornotemark[1]
\email{sunghobae@snu.ac.kr}
\affiliation{%
  \institution{Seoul National University}
  \city{Seoul}
  \country{Republic of Korea}
}

\author{Dohun Lee}
\email{dohun754@snu.ac.kr}
\affiliation{%
  \institution{Seoul National University}
  \city{Seoul}
  \country{Republic of Korea}
}

\author{Jaeeun Seo}
\email{jaeeunseo@snu.ac.kr}
\affiliation{%
  \institution{Seoul National University}
  \city{Seoul}
  \country{Republic of Korea}
}

\author{Jeehong Kim}
\email{williamkim10@snu.ac.kr}
\affiliation{%
  \institution{Seoul National University}
  \city{Seoul}
  \country{Republic of Korea}
}

\author{Wonhee Lee}
\email{weelon@kriso.re.kr}
\affiliation{%
  \institution{KRISO}
  \city{Daejeon}
  \country{Republic of Korea}
}

\author{Hyunwoo Park}
\authornote{Corresponding author.}
\email{hyunwoopark@snu.ac.kr}
\affiliation{%
  \institution{Seoul National University}
  \city{Seoul}
  \country{Republic of Korea}
}

\renewcommand{\shortauthors}{Youngseok Hwang et al.}

\begin{abstract}
Maritime anomaly detection is essential for ensuring maritime safety, security, and efficient traffic management at sea, with Automatic Identification System (AIS) data serving as a primary data source. Despite its importance, most publicly available AIS datasets lack predefined anomaly labels, forcing prior studies to rely on either distribution-based rarity or domain rule/expert-assisted labeling. These approaches, however, face fundamental limitations: statistical rarity often fails to reflect practically critical events, while expert-based labeling is costly, subjective, and difficult to scale. Moreover, both paradigms tend to overlook interaction-driven hazards such as near-miss approaches between vessels. To address these challenges, we propose an equation-grounded anomaly taxonomy that is implementable under a limited AIS observation schema and extensible to other AIS datasets. Specifically, the taxonomy defines three anomaly types: unexpected AIS activity (A1), route deviation (A2), and close approach (A3), covering both single-vessel and inter-vessel anomalies. Building on this taxonomy, we introduce a unified \emph{score--synthesize--label} pipeline that produces LLM-guided plausibility scores, uses them to synthesize anomalies, and assigns timestamp-level labels. To rigorously assess detection performance, we further design benchmark evaluation settings that account for variations in temporal-window length and anomaly-type composition, and evaluate a broad range of time-series models and anomaly detection models. Together, these contributions provide a systematic basis for evaluating maritime anomaly detection methods across different anomaly types. Our code is available at \url{https://github.com/snudial/open-maritime-anomaly-detection}.
\end{abstract}



\maketitle

\section{Introduction} \label{sec:introduction}

\begin{figure}[t]
    \includegraphics[width=\columnwidth]{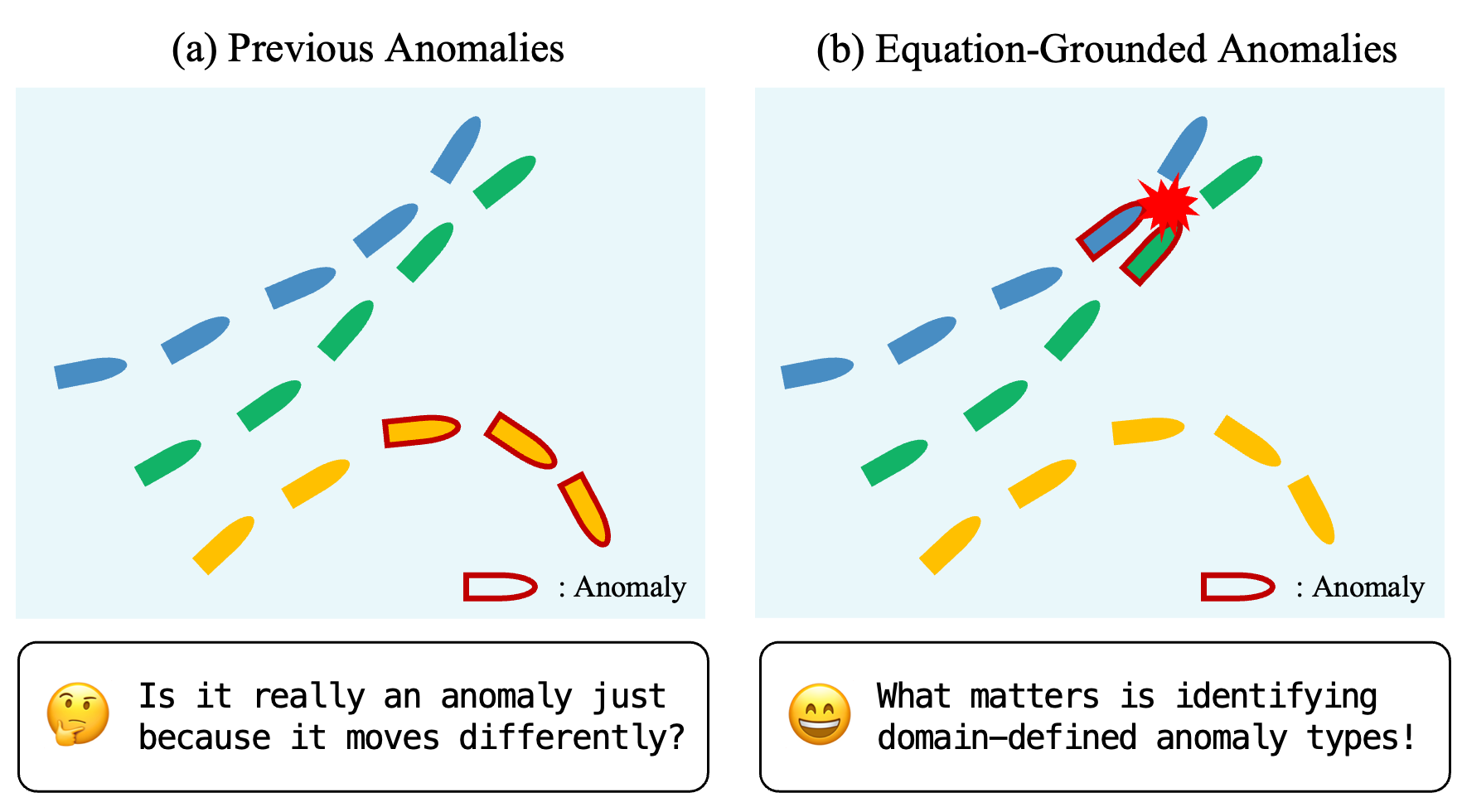}
    \Description{Motivating example comparing traditional anomaly classification and the proposed equation-grounded anomaly taxonomy. The left panel asks whether a vessel is really an anomaly just because it moves differently, while the right panel emphasizes that what matters is identifying domain-defined anomaly types.}
    \caption{Motivation and Overview. Comparison between previous and equation-grounded anomaly definitions on AIS data. Previous approaches often label deviations as anomalies, while equation-grounded definitions focus on behaviors grounded in domain knowledge.}
    \label{fig:motivation_overview}
\end{figure}

Maritime anomaly detection is a pivotal task highlighting the intersection of maritime safety, security, and traffic management \cite{may2020challenges, xie2024anomaly, sidibe2017study}. AIS data have been widely used as one of the most common sources for this purpose. However, most of the publicly available AIS datasets do not provide predefined anomaly labels. As a result, prior studies have typically defined abnormal behaviors based on either (i) distribution-based criteria or (ii) rule-based heuristics derived from current maritime operation procedures and regulations. The former often fails to capture practically critical events such as near-miss situations or abnormal maneuvers, since statistical rarity does not always indicate real-world importance. The latter requires rule-based or expert knowledge that is difficult to obtain, costly to maintain, and impractical to generalize. In publicly available AIS datasets, most observed trajectories reflect routine and successfully completed vessel operations rather than hazardous events. Consequently, assigning anomaly labels solely based on statistical sparsity can result in a significant disconnect from real-world maritime safety risks (Figure~\ref{fig:motivation_overview}).

Building upon our prior work on spatio-temporal graph representations and benchmarking for maritime anomaly detection \cite{kim2025spatio}, we formulate an equation-grounded anomaly taxonomy that remains implementable under a limited AIS observation schema while extending naturally to other AIS datasets. Specifically, we define three anomaly types inspired by prior maritime anomaly studies \cite{lane2010maritime}: (i) unexpected AIS activity (A1), (ii) route deviation (A2), and (iii) close approach (A3). A1 and A2 correspond to single-vessel anomalies, while A3 represents an inter-vessel interaction anomaly. Unexpected AIS activity captures sensor-induced positional discontinuities without accompanying kinematic changes. Route deviation models abrupt variations in course over ground (COG) and speed over ground (SOG) occurring over consecutive timestamps. Close approach is defined as a near-miss event in which the relative motion between a pair of vessels significantly increases collision risk within a specific temporal window. 

Based on this taxonomy, we propose a unified pipeline (Figure~\ref{fig:overall_framework}) following a \emph{score--synthesize--label} paradigm. For each anomaly type, we first specify explicit equation-grounded conditions that formalize the target abnormal behavior. We then synthesize anomaly data so that the resulting samples satisfy these conditions, and finally assign anomaly labels at both the segment level and the per-timestamp level according to well-defined temporal windows. In our framework, the LLM serves solely as a constrained scorer that selects the temporal window and scenario-specific parameters for anomaly generation, while all anomalies and labels are derived from the equation-grounded definitions.

To the best of our knowledge, this work is among the first to systematically formalize maritime domain knowledge into an equation-grounded anomaly taxonomy and to use an LLM as a constrained scorer for generating data-driven and kinematically consistent maritime anomalies on public AIS datasets.
Finally, we design benchmark evaluation settings that account for the differing modeling complexities across anomaly types, including variations in temporal-window length and anomaly-type composition. We evaluate a broad range of models, spanning traditional anomaly detection methods, dedicated time-series anomaly detection models, and time-series forecasting models adapted to anomaly detection through per-timestamp prediction-error scoring. We then analyze detection performance separately for each anomaly type (A1, A2, and A3).

Overall, our main contributions are as follows:
\begin{itemize}[topsep=2pt, partopsep=1pt, itemsep=1pt, parsep=1pt]
    \item We formulate maritime anomalies in vessel trajectories as an equation-grounded taxonomy, defining three distinct and operationalizable anomaly types.
    \item We propose a unified \emph{score--synthesize--label} framework with an LLM-based scorer for anomaly synthesis and labeling, which is applicable to other AIS datasets.
    \item We construct a standardized benchmark on a public AIS dataset and perform anomaly-type-specific evaluations and analyses across diverse baseline models.
\end{itemize}
\section{Related Work} \label{sec:related}
\subsection{Maritime Anomaly Detection Methods} \label{subsec:maritimeadm}
As maritime traffic density continues to grow, situation awareness and decision support in complex maritime environments become increasingly critical \cite{tian2025knowledge}. Maritime anomaly detection lies at the core of this effort, encompassing the identification of abnormal vessel behaviors, hazardous situations such as collisions and near-miss encounters, and even the principled construction of anomaly definitions \cite{xie2024anomaly, sidibe2017study}. Limitations of traditional approaches have motivated a shift toward ML/DL-based methods, spanning from conventional classifying algorithms to complex neural architectures \cite{yang2024harnessing, ma2025ais}. However, a fundamental challenge persists: the lack of publicly available AIS datasets with explicit anomaly labels. To overcome this, recent studies \cite{liang2024unsupervised, xu2025anomaly} have adopted unsupervised strategies based on trajectory distributions. For instance, \textsc{STAD} \cite{li2024stad} identifies anomalies through probabilistic models for unsupervised trajectory labeling.
However, such approaches often lack domain-specific consistency. Alternative approaches rely on expert-defined labels \cite{zhang2015method, xu2025anomaly}, which are not only costly but also prone to subjectivity. Consequently, existing evaluation processes are considered arbitrary and labor-intensive.

\subsection{Challenges in Defining Maritime Anomalies} \label{subsec:challenges}
Despite extensive work on detection models, a physically grounded and context-aware definition of anomalies remains an open challenge \cite{lane2010maritime, ribeiro2023ais, xie2024anomaly, masek2021open}. Specifically, existing definitions rarely account for inter-vessel dynamics, overlooking hazardous interactions that arise between vessels. For a detection model to truly understand and identify anomalies in an operationally meaningful way, it requires labels rooted in domain knowledge rather than mere statistical rarity. In this context, the maritime anomaly taxonomy proposed by Lane et al. \cite{lane2010maritime} has served as a foundational categorization scheme in AIS-based monitoring and threat assessment. The taxonomy identifies five anomalous ship behaviors: (i) deviation from standard routes, (ii) unexpected AIS activity, (iii) unexpected port arrival, (iv) close approach between vessels, and (v) entry into restricted or abnormal regions. Moreover, although the number of vessels involved is a key determinant of the nature and risk of maritime anomalies \cite{wen2022dynamic}, prior works \cite{riveiro2018maritime, liu2024ais, qi2026vessel} have not systematically distinguished anomalies from the perspective of vessel interaction. 
To address these gaps, we redefine the taxonomy of Lane et al. \cite{lane2010maritime} by explicitly categorizing anomalies according to the number of vessels involved, distinguishing between single-vessel anomalies and interaction-driven anomalies. We further formalize each anomaly type through equation-grounded definitions and apply them to a publicly available AIS dataset, thereby providing both a reproducible benchmark and an expandable labeling framework.

\subsection{LLM-based Synthetic Data Generation}

Synthetic data generation has emerged as a promising approach for addressing data scarcity when reliable labels are limited, improving data coverage and model robustness~\cite{havrilla2024surveying}. Recent studies have explored LLMs for synthetic data generation in tabular~\cite{kim2024epic, yang2024language} and time-series domains~\cite{rousseau2025forging, long-etal-2024-llms}, leveraging their ability to encode contextual knowledge and generate diverse scenarios beyond traditional rule-based approaches. However, directly generating time-series trajectories remains challenging in physical domains, where synthetic samples must satisfy kinematic and domain-specific constraints. In the maritime domain, \textsc{OceanGPT}~\cite{bi2024oceangpt}, LLM-MPC~\cite{zeng2026large}, and recent LLM-assisted vessel anomaly detection work~\cite{chen2025msce} show that LLMs can provide useful domain-specific and contextual priors. In anomaly detection, real anomalous events are rare and labeling is costly~\cite{blazquez2021review}, making synthetic anomaly construction a practical alternative for controlled evaluation. For AIS trajectories, however, such anomalies must remain physically and operationally consistent. Therefore, rather than directly synthesizing AIS trajectories, we employ the LLM as a constrained scoring module that guides anomaly placement and severity under predefined equation-grounded anomaly definitions.
\begin{figure*}[t]
    \centering
    \includegraphics[width=\textwidth]{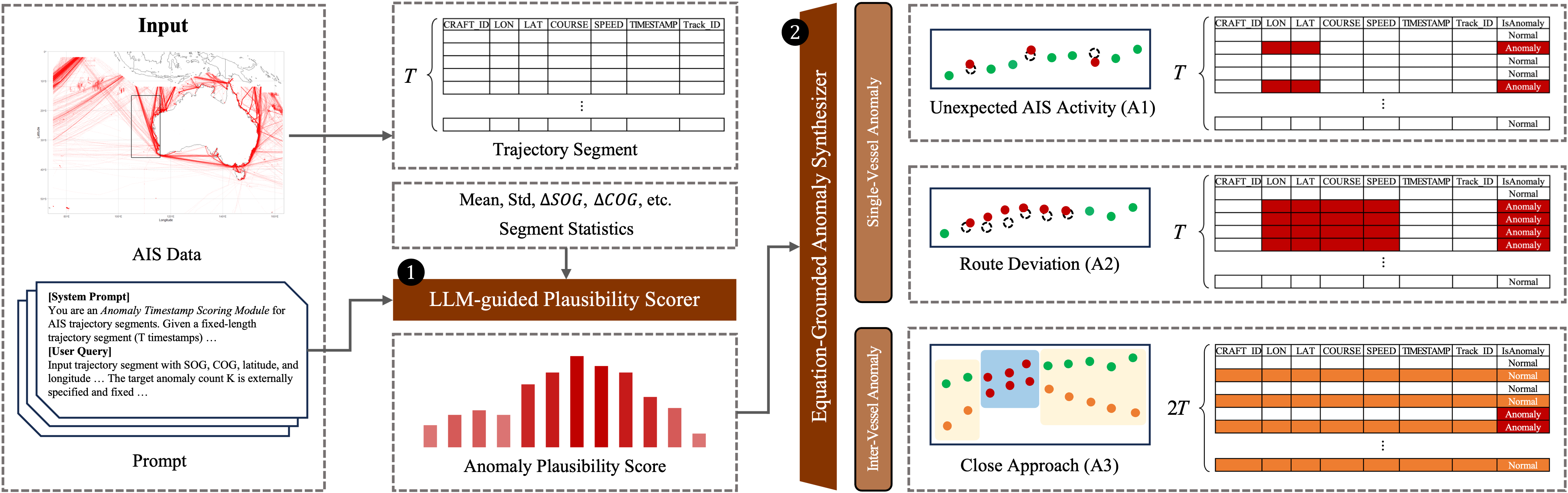}
    \Description{Pipeline diagram showing the score-synthesize-label framework with LLM-based scorer, equation-grounded synthesizer, and labeling for A1, A2, A3 anomaly types.}
    \caption{Overview of the framework. We propose a pipeline for synthesizing and labeling maritime anomalies in AIS data. An LLM-based scorer produces per-timestamp anomaly scores, which guide the placement and magnitude of synthesized anomalies. An equation-grounded synthesizer then generates anomalous values and labels for A1, A2, A3. Single-vessel anomalies span $T$ rows, while the inter-vessel anomaly forms paired samples with $2T$ rows.}
    \label{fig:overall_framework}
\end{figure*}

\section{Preliminaries} \label{sec:preliminaries}

\subsection{Notation}
We model each vessel trajectory as a multivariate time series of AIS states. Let a full trajectory be $\mathcal{X} = \{x_{0}, \ldots, x_{N-1}\}$, which we segment into non-overlapping windows of fixed length $T$, yielding segments $\mathcal{R}^{(i)}=\{x^{(i)}_{t}\}_{t=0}^{T-1}$. Each AIS state vector $x^{(i)}_{t} = (\mathbf{p}^{(i)}_{t}, \mathbf{v}^{(i)}_{t})$ is decomposed into positional $\mathbf{p}^{(i)}_{t} = (\mathrm{LAT}^{(i)}_{t}, \mathrm{LON}^{(i)}_{t})$ and kinematic $\mathbf{v}^{(i)}_{t} = (\mathrm{COG}^{(i)}_{t}, \mathrm{SOG}^{(i)}_{t})$ components, where $\mathbf{p}^{(i)}_{t}$ denotes latitude and longitude, and $\mathbf{v}^{(i)}_{t}$ denotes course and speed over ground. We report results with $T\in\{12,\ 24\}$ unless otherwise stated, excluding the first and last $T$ observations to avoid implausible boundary cases.

For anomaly synthesis, we introduce per-timestamp plausibility scores of vessel $i$ at time $t$, denoted by $a^{(i)}_{t}$, produced by an LLM-guided plausibility scorer (Section~\ref{sec:methodology}). For each synthesis run, we assign per-timestamp anomaly labels $y^{(i)}_{t}\in\{\mathrm{0}, \mathrm{1}\}$ over each trajectory segment, given a target anomaly type $A\in\{\mathrm{A1}, \mathrm{A2}, \mathrm{A3}\}$. For inter-vessel anomaly, we adopt standard maritime terminology: the own ship $o$ refers to the observed vessel and the target ship $v$ denotes a synthetically generated counterpart. Collision risk is characterized by the closest point of approach (CPA) concept under COLREGs~\cite{COLREG1972}, quantified by the distance and time to CPA (DCPA and TCPA, respectively).

\subsection{Problem Formulation} \label{subsec:problemformulation}
We define maritime anomalies into three types and formalize them using equation-grounded criteria.
We further address the problem of constructing maritime anomaly detection tasks that are suitable for learning and evaluation by systematically labeling the anomaly types defined in Table~\ref{tab:anomaly_types} from AIS data without predefined anomaly labels.

\begin{table}[!h]
\centering
\caption{Maritime anomaly types by vessel scope}
\label{tab:anomaly_types}
\resizebox{\columnwidth}{!}{
\begin{tabular}{lp{2.8cm}p{4cm}}
\toprule
\textbf{Vessel Scope} & \textbf{Anomaly Type} & \textbf{Description} \\
\midrule
\multirow{4.5}{*}{\textbf{Single-vessel}} & Unexpected AIS \newline Activity (A1) & Position spike without commensurate kinematic change \\
\cmidrule(lr){2-3}
 & \multirow{2}{*}{Route Deviation (A2)} & Abrupt maneuver with abnormal kinematic variations \\
\midrule
\multirow{2}{*}{\textbf{Inter-vessel}} & \multirow{2}{*}{Close Approach (A3)} & Interaction-driven near-miss event \\
\bottomrule
\end{tabular}
}
\end{table}

We consider three anomaly types (A1--A3) that can be consistently synthesized, labeled, and evaluated under realistic AIS observation constraints.
Unexpected AIS activity (A1) captures position spikes in latitude or longitude without commensurate changes in motion states such as SOG or COG, and is interpreted as a sensor-induced anomaly. Route deviation (A2) captures kinematics-driven abrupt maneuvers, characterized by short-window anomalies in SOG/COG that induce physically plausible but abnormal motion. Close approach (A3) captures interaction-driven near-miss events between vessels. Since public AIS data do not provide paired near-miss labels, we introduce a virtual counterpart trajectory constructed relative to the original trajectory, enabling reproducible synthesis and labeling.

We use an LLM as a plausibility scorer that reflects the semantic and contextual information of the AIS data and the predefined anomaly types. All anomaly labels are assigned deterministically based on predefined equations, and the LLM does not directly generate anomaly values or labels. Algorithm~\ref{alg:eq_grounded_labeling} summarizes the trajectory segmentation and equation-grounded labeling pipeline.

\begin{algorithm}[htbp]
\caption{Equation-grounded AIS anomaly labeling}
\label{alg:eq_grounded_labeling}
\small
\begin{algorithmic}[1]
\Require AIS trajectory set $\mathcal{D}$, window length $T$, prompt $\mathcal{P}$, target anomaly type $A \in \{\mathrm{A1}, \mathrm{A2}, \mathrm{A3}\}$
\Ensure Labeled samples $\{(\tilde{\mathcal{R}}^{(i)},\mathbf{y}^{(i)})\}$

\ForAll{$\mathcal{X}\in\mathcal{D}$}
    \State Segment $\mathcal{X}$ into fixed-length windows $\{\mathcal{R}^{(i)}\}$
    \ForAll{$\mathcal{R}^{(i)}$}
        \State Compute segment statistics $\mathbf{s}^{(i)} \leftarrow \mathrm{Stats}(\mathcal{R}^{(i)})$
        \State Compute scores $\mathbf{a}^{(i)} \leftarrow \circled{1}\,\mathrm{LLMScorer}(\mathcal{R}^{(i)},\mathbf{s}^{(i)},\mathcal{P},A)$
        \If{$A \in \{\mathrm{A1}, \mathrm{A2}\}$}
            \State $(\tilde{\mathcal{R}}^{(i)},\mathbf{y}^{(i)}) 
            \leftarrow \circled{2}\,\mathrm{AnomalySynthesizer}(\mathcal{R}^{(i)},\mathbf{a}^{(i)},A)$
            \State \hspace{\algorithmicindent}$\triangleright\ \mathbf{y}^{(i)}\in\{0,1\}^{T}$
        \Else
            \State $(\tilde{\mathcal{R}}^{(i)},\mathbf{y}^{(i)}) 
            \leftarrow \circled{2}\,\mathrm{AnomalySynthesizer}(\mathcal{R}^{(i)},\mathbf{a}^{(i)},A)$
            \State \hspace{\algorithmicindent}$\triangleright$ constructs paired segment
            \State \hspace{\algorithmicindent} \hspace{\algorithmicindent}$\tilde{\mathcal{R}}^{(i)}=[\mathcal{R}^{(i)};\mathcal{V}^{(i)}],\ 
            \mathbf{y}^{(i)}\in\{0,1\}^{2T}$
        \EndIf
        \State Append $(\tilde{\mathcal{R}}^{(i)},\mathbf{y}^{(i)})$ to output set
    \EndFor
\EndFor
\State \Return labeled samples $\{(\tilde{\mathcal{R}}^{(i)},\mathbf{y}^{(i)})\}$
\end{algorithmic}
\end{algorithm}

\section{Methodology} \label{sec:methodology}
In this section, we describe how the LLM guides anomaly synthesis under controlled and equation-grounded constraints. Figure~\ref{fig:overall_framework} illustrates the overall framework.

We provide the LLM with the trajectory segment $\mathcal{R}^{(i)}$, kinematic summary statistics $\mathbf{s}^{(i)}$, and a predefined prompt $\mathcal{P}$ that encodes dataset descriptions and anomaly definitions. The resulting output is a per-timestamp anomaly score vector: 
\begin{equation}
    \mathbf{a}^{(i)} = f_{\mathrm{LLM}}\!\left(\mathcal{R}^{(i)}, \mathbf{s}^{(i)}, \mathcal{P}\right) \in [0, 1]^{T}
    \label{eq:llm_score_def}
\end{equation} Each element ${a}^{(i)}_{t}$ represents the semantic plausibility of synthesizing an anomaly at timestamp $t$ within the segment $\mathcal{R}^{(i)}$.
Importantly, ${a}^{(i)}_{t}$ is not used to directly generate trajectory coordinates. Instead, it is used exclusively for (i) determining synthesis timestamps and (ii) scaling the severity of the anomaly in the subsequent synthesis stage. The rationale for this approach lies in the challenge of capturing the contextual and semantic nuances of diverse trajectory data. Moreover, defining explicit rules for where and how severely to inject anomalies into normal data is fundamentally impossible, as there is no ground truth in a counterfactual injection setting. Due to the uncalibrated nature of LLM outputs \cite{elazar-etal-2021-measuring}, we apply min-max normalization to capture the relative semantic plausibility within the segment. 

We use Qwen3-8B \cite{yang2025qwen3} as the LLM backbone, which enables reproducible plausibility scoring with efficient computation suitable for large-scale AIS trajectory processing.

\subsection{Single-vessel Anomalies} \label{subsec:single_vessel_real}

We generate the anomalous trajectory $\tilde{R}^{(i)}$ by perturbing the observed segment $R^{(i)} =\{\mathbf{x}_{i,t}\}_{t=0}^{T-1}$ based on an anomaly score profile $\mathbf{a}^{(i)}$. To control the anomaly intensity, we select the top-$K$ timestamps from $\mathbf{a}^{(i)}$ corresponding to fixed ratios $K/T \in \{0.25, 0.5, 0.75\}$ and semantically modulate the perturbation magnitude at each selected step.

\paragraph{\textbf{A1: Unexpected AIS Activity.}}
Unexpected AIS Activity represents position-driven anomalies where reported coordinates $\mathbf{p}^{(i)}_t$ exhibit abrupt cross-track deviations without corresponding updates in the kinematic vector $\mathbf{v}^{(i)}_t$. 
To quantify the geometric variation of the trajectory, we employ the Perpendicular Euclidean Distance (PED). We define $\mu_{\text{PED}}^{(i)}$ as the mean PED computed over all valid triplets for a trajectory segment, treating it as the average noise level of the trajectory:
\begin{equation}
\mu_{\text{PED}}^{(i)} = \frac{1}{T-2} \sum_{k=1}^{T-2} \text{PED}\left(\mathbf{p}^{(i)}_{k-1},\, \mathbf{p}^{(i)}_{k},\, \mathbf{p}^{(i)}_{k+1}\right)
\end{equation}
To ensure the generated perturbations are clearly distinguishable from average noise, we set the scaling parameter $\theta_p = 3$. For a selected timestamp $t$, we synthesize the perturbed position $\tilde{\mathbf{p}}^{(i)}_t$ by displacing the point perpendicularly to the trajectory path, ensuring it satisfies:
\begin{equation}
\text{PED}(\mathbf{p}^{(i)}_{t-1}, \tilde{\mathbf{p}}^{(i)}_t, \mathbf{p}^{(i)}_{t+1}) \ge \theta_p \cdot (1 + a^{(i)}_t) \cdot \mu_{\text{PED}}^{(i)}
\end{equation}

\paragraph{\textbf{A2: Route Deviation.}}
Route Deviation represents kinematics-driven anomalies characterized by abrupt variations in speed ($\text{SOG}$) and heading ($\text{COG}$) that propagate over time. Unlike the point-wise perturbation of A1, A2 anomalies are synthesized over a consecutive temporal window to reflect the sequential dependencies inherent in vessel motion. 

We first define the discrete changes in speed and heading between consecutive timestamps as $|\Delta \text{SOG}^{(i)}_t|$ and $|\Delta \text{COG}^{(i)}_t|$, respectively. We calculate the baseline variations $\mu^{(i)}_{\Delta \text{SOG}}$ and $\mu^{(i)}_{\Delta \text{COG}}$ by averaging these changes over a window $T$:

\begin{equation}
\mu_{\Delta \text{SOG}}^{(i)} = \frac{1}{T-1}\sum_{k=1}^{T-1} |\Delta \text{SOG}^{(i)}_k|, \;
\mu_{\Delta \text{COG}}^{(i)} = \frac{1}{T-1}\sum_{k=1}^{T-1} |\Delta \text{COG}^{(i)}_k|.
\end{equation}

Starting from the last normal state $\mathbf{x}_{i,t-1}$, the anomalous trajectory is generated by perturbing the original states and accumulating them recursively:
\begin{equation}
\begin{aligned}
\tilde{\text{SOG}}^{(i)}_t &= \text{SOG}^{(i)}_{t} + \theta_v \cdot (1 + a^{(i)}_t) \cdot \mu_{\Delta \text{SOG}}^{(i)}, \\
\tilde{\text{COG}}^{(i)}_t &= \text{COG}^{(i)}_{t} + \theta_h \cdot (1 + a^{(i)}_t) \cdot \mu_{\Delta \text{COG}}^{(i)}, \\
\tilde{\mathbf{p}}^{(i)}_{t+1} &= \tilde{\mathbf{p}}^{(i)}_{t} + \tilde{\text{SOG}}^{(i)}_t \cdot \Delta t \cdot \begin{bmatrix} \cos \tilde{\text{COG}}^{(i)}_t \\ \sin \tilde{\text{COG}}^{(i)}_t \end{bmatrix}.
\end{aligned}
\end{equation}
This recursive construction ensures that the synthesized trajectory maintains temporal continuity, while allowing the anomalous behavior to gradually deviate from the original voyage patterns. In this study, the scaling parameters $\theta_v$ and $\theta_h$ are set to $2$ to facilitate a progressive deviation. Furthermore, to ensure the vessel eventually returns to the original route, the anomaly window of size $K$ is divided into two phases: an injection phase (the first $2/3$) where the deviation accumulates, and a recovery phase (the remaining $1/3$) where the trajectory smoothly converges back to the ground truth.

\paragraph{\textbf{Labeling rule.}}
We assign a binary label to each timestamp within the synthesized segment $\tilde{\mathcal{R}}^{(i)}$. We assign $y_t^{(i)} = 1$ to the timestamps where the anomaly is injected, and $y_t^{(i)} = 0$ to all remaining timestamps. The final labeled sample is returned as $(\tilde{\mathcal{R}}^{(i)}, \mathbf{y}^{(i)})$.

\begin{figure}[t]
    \centering
    \includegraphics[width=\columnwidth]{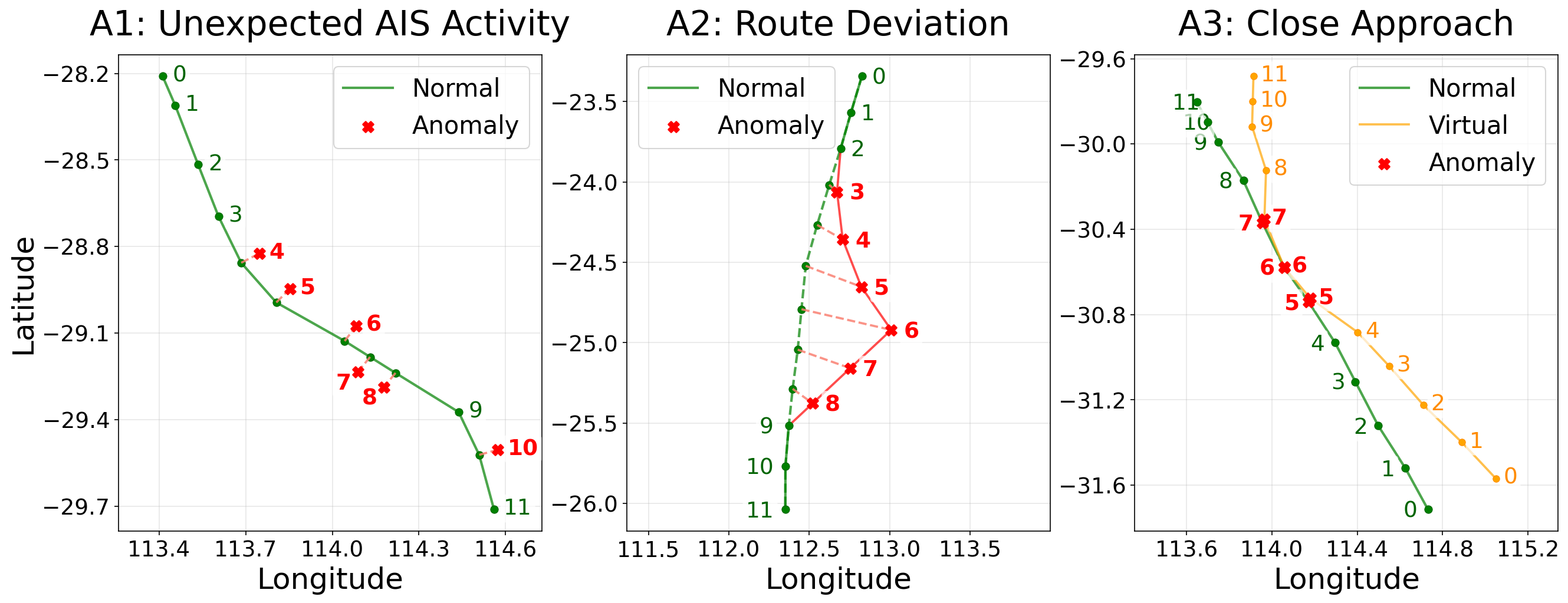}
    \Description{Three latitude-longitude trajectory plots showing synthesized anomaly examples. A1 Unexpected AIS Activity: a single vessel trajectory with scattered anomalous timestamps marked in red along the path. A2 Route Deviation: original trajectory in green with deviated positions connected by dashed red lines diverging laterally. A3 Close Approach: two vessel trajectories, one green and one orange virtual vessel, converging at timestamps 5 through 7 marked as anomalous.}
    \caption{Equation-grounded anomaly examples for three anomaly types (A1--A3)}
    \label{fig:equation_grounded_examples}
\end{figure}

\subsection{Inter-Vessel Anomaly} \label{subsec:inter_vessel_real}

Close Approach is an inter-vessel anomaly synthesized by constructing a virtual vessel segment $\mathcal{V}^{(i)}$ and pairing it with an observed segment $\mathcal{R}^{(i)}$. Importantly, A3 does not modify the own-ship trajectory. The interaction is synthesized solely through the construction of $\mathcal{V}^{(i)}$.

\paragraph{\textbf{Virtual Vessel Trajectory Generation.}}

Given the per-timestamp anomaly scores $\mathbf{a}_t$, we select the anchor index as $k = \arg\max_j \mathbf{a}_j$, where ties are broken uniformly at random, and define the length-$3$ anomalous window as $\{k-1, k, k+1\}$. We represent the own-ship positions $\mathbf{p}^{(o)}_t$ in a local East-North-Up (ENU) frame anchored at timestamp $k$, i.e., a local east–north Cartesian coordinate system centered at the anchor point.

We synthesize a virtual vessel by laterally offsetting the own-ship trajectory:
\begin{equation}
\mathbf{p}^{(v)}_{t}=\mathbf{p}^{(o)}_{t}+ d_t\,\mathbf{n}_{t},
\qquad t=0,\dots,T{-}1,
\label{eq:a3_pos_offset}
\end{equation}
where $d_t\ge 0$ controls the magnitude of the pairwise separation and $\mathbf{n}_t\in\mathbb{R}^2$ is a unit lateral direction. At the anchor point, we assign a virtual velocity $\mathbf{v}^{(v)}$ according to predefined speed statistics which leads the relative velocity to be defined as

\(
\mathbf{u} = \mathbf{v}^{(v)} - \mathbf{v}^{(o)}.
\)
Based on this relative motion, we select a reference lateral unit normal $\mathbf{n}_{k}$ perpendicular to $\mathbf{u}$, with a consistent left/right orientation with respect to the own-ship heading.

We then specify a minimum target separation $\widetilde{D}$ as a surrogate for the DCPA concept and place the virtual vessel's anchor point at a lateral offset of $\widetilde{D}$ along $\mathbf{n}_{k}$. In practice, $\widetilde{D}$ is sampled from the \textbf{DANGEROUS} range in our risk taxonomy (Table~\ref{tab:risk_classification}), i.e., $\widetilde{D}\in[0.1,\,0.3]$~NM, guided by the anomaly plausibility scores.

\begin{table}[h]
\centering
\scriptsize
\caption{Close Approach Risk Level Classification}
\label{tab:risk_classification}
\resizebox{\columnwidth}{!}{
\begin{tabular}{lccc}
    \toprule
    \textbf{Level} & \textbf{DCPA} (NM) & \textbf{TCPA} (min) & \textbf{Reference} \\
    \midrule
    CRITICAL & $< 0.1$ & $< 3$ & \cite{yoo2018near} \\
    \textbf{DANGEROUS} & $0.1$--$0.3$ & $3$--$5$ & \cite{IMO1995A823, coldwell1983marine, sui2023real} \\
    CLOSE & $0.3$--$0.5$ & $5$--$10$ & \cite{goodwin1975statistical} \\
    WATCH & $0.5$--$1.0$ & $10$--$15$ & \cite{goodwin1975statistical, ha2021quantitative}\\
    MONITOR & $\geq 1.0$ & $\geq 15$ & \cite{IMO1995A823} \\ 
    \bottomrule
\end{tabular}
}
\end{table}

The taxonomy described in Table \ref{tab:risk_classification} is calculated using the metadata of vessel directory from International Maritime Organization, focusing on the cargo ships and tankers. DCPA and TCPA are dependent on average vessel lengths and heights (of the masts).  
Given this anchor configuration, the time-varying separation distance $d_t$ is constructed to follow a CPA-inspired baseline profile~\cite{COLREG1972}, ensuring that the minimum separation occurs at $k$ and increases smoothly away from the anchor point.

To avoid a trivial parallel translation of the own-ship trajectory, the lateral direction $\mathbf{n}_t$ is allowed to vary over time by rotating $\mathbf{n}_k$ according to a score-derived heading profile $\theta(t)$, and the resulting lateral offset is given by $\mathbf{r}_t = d_t \mathbf{n}_t$.

Finally, the full virtual vessel trajectory is obtained as
$\mathbf{p}^{(v)}_t = \mathbf{p}^{(o)}_t + \mathbf{r}_t$, and its kinematic states are recovered deterministically through finite differences in the ENU frame, ensuring consistency between position, speed, and course.

\paragraph{\textbf{Labeling rule.}}
We limit each anomaly event to a three-timestamp window to filter out normal co-traveling behaviors. Accordingly, we assign
\(
y_t^{(i)} = 1
\)
for
\(
t \in \{k-1, k, k+1\}
\)
and
\(
y_t^{(i)} = 0
\)
otherwise.
For A3, the labeled sample is formed as a paired segment 
\(
\tilde{\mathcal{R}}^{(i)} = [\mathcal{R}^{(i)}; \mathcal{V}^{(i)}],
\)
and the same per-timestamp labels are applied to both the own-ship and virtual-ship rows, since the anomaly is defined as a pairwise event. 
The labeled sample $(\tilde{\mathcal{R}}^{(i)}, \mathbf{y}^{(i)})$ is returned as the result. This design constrains A3 anomalies to short-term proximity events of at most three hours, consistent with the CPA/DCPA-based interpretation of collision risk under COLREGs~\cite{COLREG1972} (Rules~7--8).

\section{LLM Scorer Validation} \label{sec:llm_validation}

\subsection{Distributional Analysis}
\begin{figure}[h]
    \centering
    \includegraphics[width=\columnwidth]{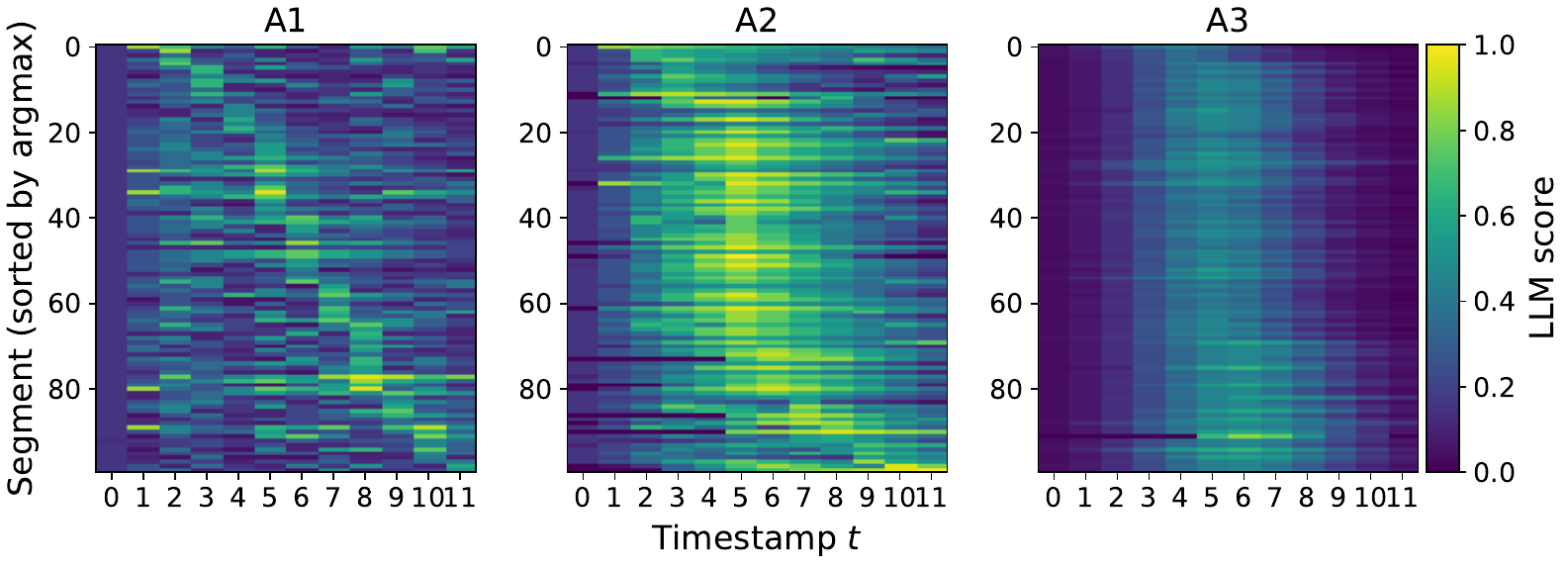}
    \caption{Distributional analysis of LLM scores across anomaly types. Each panel shows per-timestamp LLM scores for 100 trajectory segments randomly selected from the $N{=}1000$ sampled segments.}
    \Description{Three heatmaps of per-timestamp LLM scores for 100 trajectory segments with T equal to 12, one panel each for anomaly types A1, A2, and A3. Rows are segments sorted by argmax and columns are timestamps 0 to 11; color encodes the score from 0 to 1. The A1 panel shows sparse isolated peaks whose positions vary across segments, the A2 panel shows broad contiguous bands of high scores, and the A3 panel shows a unimodal profile concentrated around mid-segment timestamps.}
    \label{fig:score_heatmap}
\end{figure}

To verify that the LLM scorer produces anomaly-type-aware guidance rather than assigning high scores to fixed timestamp positions, we sample $N=1000$ normal trajectory segments with $T=12$ and query the LLM once per segment for each anomaly type. As shown in Figure~\ref{fig:score_heatmap}, which visualizes the score profiles of 100 randomly selected segments, A1 produces sparse peaks whose positions vary across segments, A2 forms broad contiguous high-score regions consistent with route-deviation anomalies, and A3 produces an anchor-centered score profile suitable for constructing a close-approach event. These patterns confirm that the scorer is compatible with the synthesis rules for each anomaly type.

\subsection{Repeatability}
\begin{figure}[h]
    \centering
    \includegraphics[width=0.8\columnwidth]{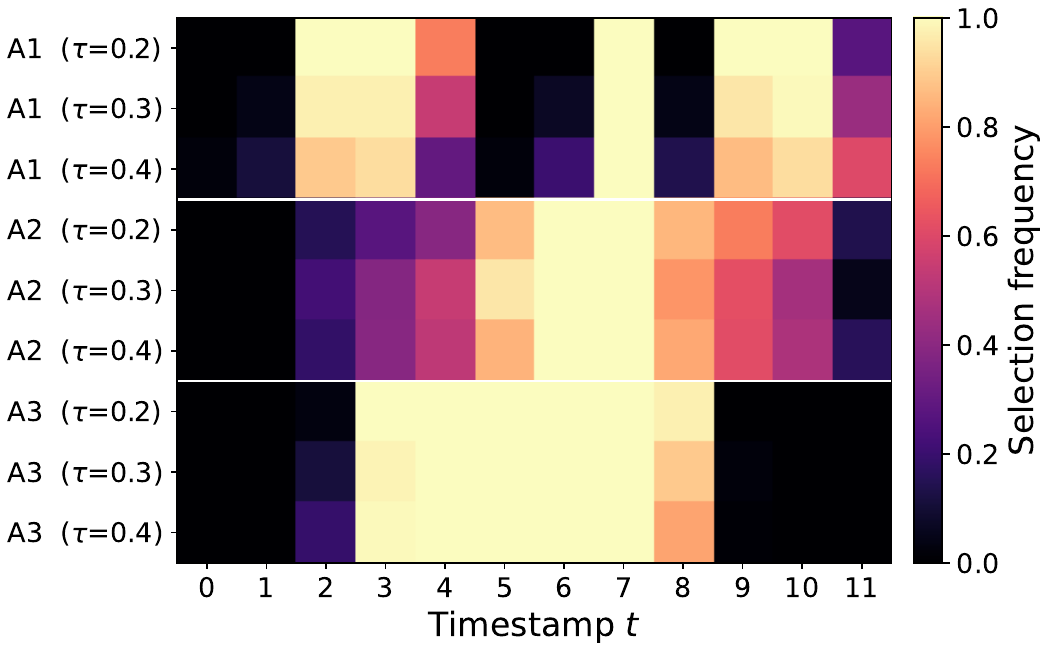}
    \caption{Decision-level repeatability of the LLM scorer. Each cell shows the selection frequency of timestamp $t$ over $R=100$ repeated queries under the same input.}
    \Description{Heatmap of timestamp selection frequencies over 100 repeated LLM queries under identical input. Rows correspond to the nine combinations of anomaly type A1, A2, or A3 with decoding temperature 0.2, 0.3, or 0.4; columns are timestamps 0 to 11; color encodes selection frequency from 0 to 1. Within each anomaly type, the high-frequency timestamps remain nearly identical across the three temperatures, indicating stable selections.}
    \label{fig:repeatability_heatmap}
\end{figure}

Since the synthesizer uses score rankings rather than exact values, we further examine whether repeated queries under the same input yield stable timestamp selections. We query the LLM $R=100$ times with varying decoding temperature $\tau \in \{0.2, 0.3, 0.4\}$ and record the top-$K$ selected timestamps. As shown in Figure~\ref{fig:repeatability_heatmap}, selection patterns remain stable across temperatures for all three anomaly types, supporting the reliable use of the LLM scorer as a guide for equation-grounded anomaly synthesis.

Together, the distributional and repeatability analyses show that the LLM scorer produces anomaly-type-aware score profiles whose timestamp selections remain stable across repeated queries and temperature settings. This supports using LLM-derived scores to guide anomaly timestamp selection and normalized severity modulation in the equation-grounded synthesizer.

Given these score profiles, we partition the set of trajectory segments and pre-assign a target anomaly type to each segment to maintain a balanced anomaly-type composition. For each assigned segment $\mathcal{R}^{(i)}$, anomalous values and row-wise labels are then generated deterministically according to the corresponding equation-grounded conditions. A1 and A2 are grouped as single-vessel anomalies because they modify the observed trajectory values, whereas A3 follows a paired inter-vessel synthesis paradigm in which the observed own-ship trajectory remains intact and a virtual counterpart is constructed. Figure~\ref{fig:equation_grounded_examples} shows representative synthesized and labeled examples for A1--A3.

\subsection{Backbone Ablation}
We evaluate three Qwen3 backbones (4B, 8B, 14B) on $n = 6{,}289$ segments per anomaly type ($T=12$) and measure pairwise agreement via Pearson~$r$ on raw scores and Jaccard similarity on top-$K$ selected timestamps. A3 scores are structurally constrained to a unimodal mid-segment profile, leaving little room for backbone-dependent variation, so we focus on A1 and A2. As shown in Table~\ref{tab:backbone_ablation}, Pearson $r$ ranges from 0.52 to 0.71, indicating that different backbones agree on which regions are suitable for anomaly placement. No single model size introduces a systematic bias, supporting the backbone-agnostic use of the LLM scorer.

\begin{table}[h]
\centering
\caption{Pairwise backbone agreement on LLM scorer outputs.}
\label{tab:backbone_ablation}
\resizebox{\columnwidth}{!}{%
\begin{tabular}{lcccc}
\toprule
Pair & A1 Pearson $r$ & A1 Top-$K$ Jaccard & A2 Pearson $r$ & A2 Top-$K$ Jaccard \\
\midrule
4B vs 8B  & $0.565$ & $0.580$ & $0.629$ & $0.612$ \\
4B vs 14B & $0.609$ & $0.607$ & $0.521$ & $0.560$ \\
8B vs 14B & $0.597$ & $0.606$ & $0.710$ & $0.656$ \\
\bottomrule
\end{tabular}%
}
\end{table}

\section{Experiments}

\subsection{Synthetic Anomaly Dataset}
We use the publicly available AIS dataset OMTAD~\cite{masek2021open}. In OMTAD, each vessel track corresponds to a navigation leg within a region of interest near the western coast of Australia. The dataset spans three years (2018--2020) and includes four vessel categories: cargo, tanker, fishing, and passenger. In this work, we conduct our analysis on cargo and tanker tracks. Fishing and passenger tracks are relatively scarce and exhibit highly irregular, task-driven movement patterns unsuitable for equation-grounded synthesis~\cite{li2024stad, liu2024ais, yu2025aisformer}. In contrast, cargo and tanker vessels typically follow more stable and repetitive route patterns, providing a better foundation for defining and injecting structurally controlled anomalies. As described in Section~\ref{sec:methodology}, we construct anomaly-labeled datasets for three anomaly types (A1, A2, and A3) by leveraging LLM-derived anomaly plausibility scores.

\begin{table}[h]
\centering
\tiny
\renewcommand{\arraystretch}{0.9}
\caption{Single-timestamp level dataset statistics at anomaly ratio of 5\%.}
\label{tab:dataset_stats_5pct}
\resizebox{\columnwidth}{!}{%
\begin{tabular}{c c c c c c}
\toprule
\multirow{2.5}{*}{$T$} & \multirow{2.5}{*}{\textbf{Type}}
& \multicolumn{2}{c}{\textbf{Train}}
& \multicolumn{2}{c}{\textbf{Test}} \\
\cmidrule(lr){3-4} \cmidrule(lr){5-6}
& & Normal & Anomaly (\textcolor{red}{\%}) & Normal & Anomaly (\textcolor{red}{\%}) \\
\midrule
\multirow{3}{*}{\textbf{12}}
& \textbf{A1} & 516{,}042 & 12{,}846 (\textcolor{red}{2.43}) & 110{,}565 & 2{,}751 (\textcolor{red}{2.43}) \\
& \textbf{A2} & 516{,}039 & 12{,}837 (\textcolor{red}{2.43}) & 110{,}574 & 2{,}754 (\textcolor{red}{2.43}) \\
& \textbf{A3} & 568{,}518 & 13{,}206 (\textcolor{red}{2.27}) & 121{,}815 & 2{,}829 (\textcolor{red}{2.27}) \\
\midrule
\multirow{3}{*}{\textbf{24}}
& \textbf{A1} & 460{,}350 & 11{,}058 (\textcolor{red}{2.35}) & 98{,}658 & 2{,}382 (\textcolor{red}{2.36}) \\
& \textbf{A2} & 460{,}338 & 11{,}070 (\textcolor{red}{2.35}) & 98{,}664 & 2{,}376 (\textcolor{red}{2.35}) \\
& \textbf{A3} & 512{,}655 & 5{,}889 (\textcolor{red}{1.14}) & 109{,}860 & 1{,}260 (\textcolor{red}{1.13}) \\
\bottomrule
\end{tabular}
}
\end{table}

For experiments, we segment each track into fixed-length sequences with $T \in \{12, 24\}$ timestamps. We then generate synthetic anomalies at four anomaly ratios (1\%, 3\%, 5\%, and 10\%), enabling systematic evaluation across different settings. 
As a representative setting, the dataset statistics under the 5\% anomaly ratio are summarized in Table~\ref{tab:dataset_stats_5pct}. The table shows how the number of anomalous timestamps varies across anomaly types (A1--A3), reflecting heterogeneity in their anomaly-specific labeling rules. Since the anomaly ratio is applied at the segment level, the resulting proportion of anomalous timestamps can be lower than the nominal ratio. Specifically, the number of timestamps selected for injection is derived by scaling the segment length $T$ by a factor chosen from $\{0.25, 0.5, 0.75\}$. We use this 5\% setting to compare equation-grounded anomalies with distribution-based anomalies.

\begin{figure}[h]
    \centering
    \includegraphics[width=\columnwidth]{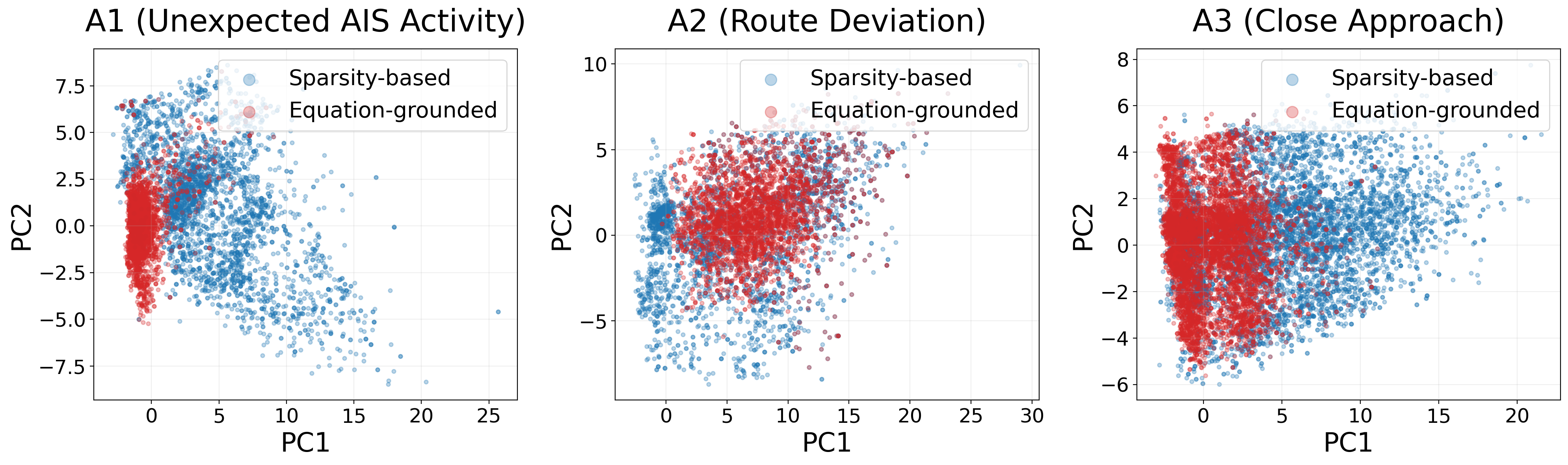}
    \Description{Three PCA scatter plots comparing sparsity-based anomalies in blue and equation-grounded anomalies in red for A1, A2, and A3. For A1 and A3, the two groups occupy largely distinct regions of the feature space, while for A2 they overlap substantially, indicating that sparsity-based selection captures route-deviation anomalies but misses most position-spike and close-approach anomalies.}
    \caption{Comparison of sparsity-based and equation-grounded anomalies in the feature space (PCA).}
    \label{fig:sparsity_comparison}
\end{figure}

Figure~\ref{fig:sparsity_comparison} visualizes the distributional differences in an embedding space constructed from representations of synthesized anomaly features, which are projected onto two dimensions via PCA. Given the same number of anomalous timestamps between sparsity-based selection and equation-grounded anomalies, A1 and A3 exhibit substantially less overlap compared to A2. 

Table~\ref{tab:recall_sparsity} reports the recall of equation-grounded anomalies as the selection proportion of sparsity-based methods increases, where recall is defined as
$
\text{Recall}
=
\frac{
\left|\, \text{sparsity-selected} \;\cap\; \text{equation-grounded anomalies} \,\right|
}{
\left|\, \text{equation-grounded anomalies} \,\right|
}. $ Even with large selection proportions, a substantial fraction of equation-grounded anomalies remains uncaptured—particularly for A1 and A3. Together, these results demonstrate that physically meaningful anomalies are not necessarily characterized by statistical rarity.

\begin{table}[h]
\centering
\caption{Recall of equation-grounded anomalies under sparsity-based selection proportions (\%) at an anomaly ratio of 5\%.}
\label{tab:recall_sparsity}
\resizebox{\columnwidth}{!}{%
\setlength{\tabcolsep}{4.5pt}
\begin{tabular}{c|c c c c c c c c c c}
\toprule
\textbf{Type} 
& \textbf{5\%} & \textbf{10\%} & \textbf{15\%} & \textbf{20\%} & \textbf{25\%}
& \textbf{30\%} & \textbf{35\%} & \textbf{40\%} & \textbf{45\%} & \textbf{50\%} \\
\midrule
\textbf{A1} 
& 3.31 & 6.01 & 8.89 & 13.23 & 18.36 & 23.84 & 28.71 & 34.14 & 38.74 & 44.40 \\
\textbf{A2} 
& 22.96 & 47.82 & 69.78 & 83.54 & 91.33 & 96.31 & 98.48 & 99.36 & 99.80 & 99.85 \\
\textbf{A3} 
& 4.42 & 10.36 & 16.57 & 23.11 & 31.12 & 39.02 & 46.81 & 54.04 & 60.06 & 65.21 \\
\bottomrule
\end{tabular}
}
\end{table}

\subsection{Evaluation Protocol} 
We perform maritime anomaly detection for all three anomaly types (A1--A3) under multiple anomaly ratio settings. We consider two sequence lengths, $T \in \{12, 24\}$, and report results separately for each anomaly type. We adopt a strict point-wise evaluation protocol that directly compares per-timestamp predictions against ground-truth labels without any point adjustment, event-level matching, or window-based tolerance. This is consistent with our synthesis procedure, which assigns anomaly labels at the individual timestamp level. We use AUROC and AUPRC as primary threshold-free metrics, alongside the best achievable F1 score, which is obtained by sweeping thresholds over the anomaly score distribution and reporting the maximum F1 across all thresholds. This allows F1 to reflect the model's upper-bound point-level discriminative capacity independent of threshold selection.

\subsection{Experimental Setup}
Our baselines span a broad range of approaches, including traditional anomaly detection methods, dedicated time-series anomaly detection models, and forecasting-based models adapted to anomaly detection. For these models, we use per-timestamp MSE prediction error as the anomaly score rather than a classifier head, as this score-based formulation is more aligned with the anomaly detection task. Detailed descriptions of baselines are provided in Appendix~\ref{app:ad}. All experiments are implemented in PyTorch and run on NVIDIA RTX 3090 and H100 GPUs.

\subsection{Experimental Results}

The experimental results are summarized in Table~\ref{tab:baseline}. Since each anomaly type exhibits distinct characteristics, we analyze and report the detection performance separately for each anomaly type. We characterize the detection challenges posed by each anomaly type across diverse model architectures.

\begin{table*}[t]
\centering
\renewcommand{\arraystretch}{1.1}
\caption{Timestamp-level anomaly detection at a 5\% anomaly ratio (AUROC / AUPRC / Best F1). \textbf{Bold} and \underline{underline} denote the best and second-best scores per metric within each setting.}
\label{tab:baseline}
\resizebox{\textwidth}{!}{%
\begin{tabular}{l|ccc|ccc|ccc|ccc|ccc|ccc}
\toprule
\multicolumn{1}{c|}{\textbf{Anomaly Type}} 
& \multicolumn{6}{c|}{\textbf{A1: Unexpected AIS Activity}} 
& \multicolumn{6}{c|}{\textbf{A2: Route Deviation}} 
& \multicolumn{6}{c}{\textbf{A3: Close Approach}} \\
\midrule
\multicolumn{1}{c|}{\textbf{Window Length}} 
& \multicolumn{3}{c|}{\textbf{$T$=12}} 
& \multicolumn{3}{c|}{\textbf{$T$=24}} 
& \multicolumn{3}{c|}{\textbf{$T$=12}} 
& \multicolumn{3}{c|}{\textbf{$T$=24}} 
& \multicolumn{3}{c|}{\textbf{$T$=12}} 
& \multicolumn{3}{c}{\textbf{$T$=24}} \\
\midrule
\multicolumn{1}{c|}{\textbf{Metric}} 
& \textbf{ROC} & \textbf{PRC} & \textbf{F1}
& \textbf{ROC} & \textbf{PRC} & \textbf{F1}
& \textbf{ROC} & \textbf{PRC} & \textbf{F1}
& \textbf{ROC} & \textbf{PRC} & \textbf{F1}
& \textbf{ROC} & \textbf{PRC} & \textbf{F1}
& \textbf{ROC} & \textbf{PRC} & \textbf{F1} \\
\midrule

\textbf{LOF} \cite{breunig2000lof}
& 50.16 & 2.41 & 4.78
& 50.13 & 2.40 & 4.69
& 83.88 & 12.63 & 20.47
& 78.10 & 10.39 & 18.79
& 55.67 & 2.77 & 5.53
& 54.39 & 1.29 & 2.71 \\

\textbf{OCSVM} \cite{scholkopf2001ocsvm}
& 49.42 & 2.31 & 4.80
& 47.33 & 2.14 & 4.63
& 79.74 & 18.13 & 27.19
& 82.88 & 20.66 & 29.68
& 49.45 & 2.20 & 4.46
& 49.49 & 1.07 & 2.31 \\

\textbf{IForest} \cite{liu2008iforest}
& 49.71 & 2.32 & 4.86
& 48.34 & 2.20 & 4.65
& 85.52 & 10.19 & 19.93
& 88.81 & 13.51 & 23.22
& 52.44 & 2.35 & 4.78
& 51.38 & 1.13 & 2.44 \\
\midrule

\textbf{LSTM-VAE} \cite{park2018lstmvae}
& 49.60 & 2.33 & 4.79
& 47.52 & 2.16 & 4.64
& 90.96 & 19.55 & 31.35
& \underline{93.36} & 23.11 & \underline{34.56}
& 64.13 & 4.67 & 11.04
& 62.38 & 1.97 & 5.11 \\

\textbf{OmniAnomaly} \cite{su2019omnianomaly}
& 45.63 & 2.12 & 4.75
& 44.13 & 1.98 & 4.61
& \underline{93.07} & 20.76 & 31.35
& \textbf{94.70} & \textbf{25.53} & \textbf{34.80}
& 54.69 & 3.00 & 5.67
& 55.59 & 1.46 & 3.41 \\

\textbf{USAD} \cite{audibert2020usad}
& 49.84 & 2.35 & 4.79
& 48.48 & 2.19 & 4.65
& 81.81 & 14.57 & 20.21
& 83.68 & 16.75 & 22.77
& 50.09 & 2.24 & 4.48
& 50.81 & 1.12 & 2.38 \\

\textbf{TranAD} \cite{tuli2022tranad}
& 49.33 & 2.36 & 4.80
& 48.41 & 2.24 & 4.66
& 72.74 & 9.65 & 14.24
& 78.42 & 11.83 & 16.54
& 50.63 & 2.30 & 4.59
& 52.11 & 1.18 & 2.51 \\

\textbf{Anomaly Transformer} \cite{xu2022anomaly}
& 51.49 & 2.48 & 4.99
& 51.75 & 2.39 & 4.89
& 54.74 & 6.94 & 16.86
& 40.56 & 4.29 & 9.39
& \textbf{75.42} & \textbf{7.88} & \textbf{16.12}
& 49.88 & 1.09 & 2.30 \\
\midrule

\textbf{LSTM} \cite{hochreiter1997lstm}
& \underline{54.42} & \underline{2.80} & 5.76
& 53.09 & \underline{2.58} & 5.19
& 92.95 & \underline{26.35} & \underline{36.60}
& 92.15 & 22.13 & 32.48
& 66.07 & 4.57 & 10.98
& \textbf{66.59} & \textbf{2.47} & \underline{6.71} \\

\textbf{VanillaTransformer} \cite{vaswani2017transformer}
& 54.34 & \underline{2.80} & \underline{5.79}
& \textbf{53.46} & \textbf{2.60} & \textbf{5.27}
& \textbf{94.39} & \textbf{28.84} & \textbf{38.85}
& 92.64 & \underline{23.60} & 33.56
& \underline{66.77} & \underline{4.84} & \underline{12.14}
& \underline{66.26} & \textbf{2.47} & \textbf{6.89} \\

\textbf{Informer} \cite{zhou2021informer}
& \textbf{54.59} & \textbf{2.83} & \textbf{5.81}
& \underline{53.25} & \textbf{2.60} & \underline{5.25}
& 91.91 & 25.45 & 35.87
& 91.84 & 22.14 & 32.27
& 65.45 & 4.49 & 10.91
& 66.13 & \underline{2.43} & 6.70 \\

\textbf{Autoformer} \cite{wu2021autoformer}
& 50.55 & 2.58 & 5.22
& 49.41 & 2.33 & 4.84
& 65.30 & 7.61 & 13.15
& 76.62 & 12.86 & 19.44
& 49.32 & 2.39 & 4.86
& 48.75 & 1.12 & 2.39 \\

\textbf{FEDformer} \cite{zhou2022fedformer}
& 50.84 & 2.61 & 5.23
& 49.53 & 2.34 & 4.87
& 58.72 & 5.50 & 9.84
& 71.34 & 11.24 & 17.38
& 48.40 & 2.33 & 4.85
& 49.14 & 1.14 & 2.41 \\

\textbf{Pyraformer} \cite{liu2022pyraformer}
& 51.34 & 2.62 & 5.32
& 51.33 & 2.47 & 4.96
& 83.70 & 12.26 & 22.47
& 88.84 & 17.60 & 28.65
& 57.46 & 3.03 & 6.59
& 58.36 & 1.55 & 3.63 \\

\textbf{PatchTST} \cite{nie2023patch}
& 46.25 & 2.31 & 5.09
& 46.73 & 2.19 & 4.76
& 86.42 & 8.93 & 19.44
& 89.03 & 10.51 & 21.98
& 54.54 & 2.62 & 6.16
& 56.60 & 1.41 & 3.39 \\

\textbf{DLinear} \cite{zeng2023dlinear}
& 48.75 & 2.48 & 5.19
& 50.34 & 2.44 & 4.92
& 66.11 & 5.28 & 12.08
& 64.94 & 4.22 & 9.62
& 63.73 & 3.37 & 8.38
& 56.56 & 1.35 & 3.38 \\

\midrule
\multicolumn{19}{l}{\textit{Note:} ROC and PRC denote AUROC and AUPRC; F1 denotes Best F1 (maximum F1 over threshold sweep).} \\
\bottomrule
\end{tabular}%
}
\end{table*}

\paragraph{\textbf{A1: Unexpected AIS Activity.}}
For A1, forecasting-based models using MSE prediction error as the anomaly score consistently outperform dedicated anomaly detection models across both window lengths. Informer achieves the strongest performance across all three metrics for $T{=}12$ (54.59, 2.83, 5.81), with VanillaTransformer and LSTM following closely, while VanillaTransformer leads for $T{=}24$ (53.46, 2.60, 5.27), with Informer ranking second. In contrast, dedicated anomaly detection models such as Anomaly Transformer, OmniAnomaly, and LSTM-VAE show near-random AUROC ($\approx$44--52), suggesting that their unsupervised reconstruction-based objectives are poorly aligned with position-only perturbations that leave kinematic signals intact. These trends reflect the defining characteristics of A1, where anomalies appear as sparse timestamp-level positional irregularities without corresponding kinematic changes. The dominant cue is a subtle, channel-specific deviation rather than a pronounced kinematic discontinuity, making A1 easily confounded with normal positional variability. Nevertheless, even top-performing models achieve only marginal gains over random (AUROC $\approx$54), indicating that position-only spikes without kinematic changes pose a fundamentally distinct detection challenge.

\paragraph{\textbf{A2: Route Deviation.}}
For A2, performance improves substantially over A1 across all models, reflecting that kinematic-coupled deviations provide clearer discriminative cues than position-only perturbations. At the 5\% anomaly ratio, VanillaTransformer achieves the strongest performance across all three metrics for $T{=}12$ (94.39, 28.84, 38.85), with LSTM ranking second in AUPRC and F1 (26.35, 36.60), indicating that forecasting-based models with prediction-error scoring effectively capture salient kinematic deviations. For $T{=}24$, OmniAnomaly achieves the strongest performance across all three metrics (94.70, 25.53, 34.80), with LSTM-VAE ranking second in AUROC and F1, suggesting that probabilistic reconstruction-based models benefit from longer context when modeling the sequential propagation of route deviations. Notably, the leading models differ between $T{=}12$ and $T{=}24$, indicating that the optimal inductive bias for A2 detection depends on the available temporal context. Overall, A2 is more amenable to detection than A1, as the anomaly mechanism produces a distinctive kinematic signature that discriminative and generative models can both exploit.

\begin{table}[t]
\centering
\caption{Comparison of single-trajectory and pairwise inter-vessel representations for A3 under F1 score. 
STGVAD is evaluated only with pairwise inputs because its graph construction requires inter-vessel relations. 
Results are from a single experimental run and may differ from Table~\ref{tab:baseline}, which reports averages over multiple runs.}
\label{tab:a3_exp}
\resizebox{\columnwidth}{!}{%
\begin{tabular}{cc|c|cc|cc}
\toprule
\multirow{3}{*}{\textbf{Ratio}} 
& \multirow{3}{*}{$T$} 
& \multicolumn{1}{c|}{\textbf{Single}} 
& \multicolumn{4}{c}{\textbf{Pairwise Inter-vessel Input}} \\
\cmidrule(lr){3-3} \cmidrule(lr){4-7}
& & \textbf{LSTM}
& \multicolumn{2}{c|}{\textbf{LSTM}} 
& \multicolumn{2}{c}{\textbf{STGVAD}} \\
\cmidrule(lr){3-3} \cmidrule(lr){4-7}
& & \textbf{Base}
& \textbf{Concat} & \textbf{Diff} 
& \textbf{Concat} & \textbf{Diff} \\
\midrule

\multirow{2}{*}{\textbf{1\%}}  
& \textbf{12} & 3.59 & 4.79 \inc{1.20} & \textbf{8.28} \inc{4.69} & 7.51 \inc{3.92} & \underline{8.14} \inc{4.55} \\
& \textbf{24} & 2.24 & 2.03 \dec{-0.21} & \underline{4.40} \inc{2.16} & 3.36 \inc{1.12} & \textbf{6.59} \inc{4.35} \\

\midrule
\multirow{2}{*}{\textbf{3\%}}  
& \textbf{12} & 9.89 & 21.04 \inc{11.15} & \textbf{31.99} \inc{22.10} & 27.20 \inc{17.31} & \underline{28.12} \inc{18.23} \\
& \textbf{24} & 6.23 & 7.58 \inc{1.35} & \underline{17.18} \inc{10.95} & 17.01 \inc{10.78} & \textbf{26.56} \inc{20.33} \\ 

\midrule
\multirow{2}{*}{\textbf{5\%}}  
& \textbf{12} & 16.13 & 32.69 \inc{16.56} & \underline{41.99} \inc{25.86} & 39.19 \inc{23.06} & \textbf{42.41} \inc{26.28} \\
& \textbf{24} & 10.82 & 13.48 \inc{2.66} & 22.76 \inc{11.94} & \underline{28.85} \inc{18.03} & \textbf{38.44} \inc{27.62} \\ 

\midrule
\multirow{2}{*}{\textbf{10\%}} 
& \textbf{12} & 29.86 & 49.66 \inc{19.80} & \underline{55.86} \inc{26.00} & 55.23 \inc{25.37} & \textbf{57.54} \inc{27.68} \\
& \textbf{24} & 19.77 & 28.83 \inc{9.06} & 39.15 \inc{19.38} & \underline{47.29} \inc{27.52} & \textbf{53.74} \inc{33.97} \\ 

\bottomrule
\end{tabular}%
}
\end{table}

\paragraph{\textbf{A3: Close Approach.}}
For A3, performance falls between A1 and A2, reflecting the inherently interaction-driven nature of the anomaly, where detection depends on modeling pairwise relative-motion dynamics rather than properties of individual trajectories. 
While A3 outperforms A1 in terms of AUROC (particularly for models such as Anomaly Transformer achieving 75.42), it remains substantially below A2, indicating that interaction-driven cues are more detectable than position-only spikes but harder to capture than kinematic deviations. At the 5\% anomaly ratio, Anomaly Transformer achieves the strongest performance across all three metrics for $T{=}12$ (75.42, 7.88, 16.12), leveraging its association discrepancy mechanism to capture interaction-induced temporal inconsistencies. For $T{=}24$, forecasting-based models dominate: LSTM leads in AUROC (66.59) and VanillaTransformer leads in F1 (6.89), suggesting that these models produce better-calibrated anomaly scores when interaction cues propagate over longer temporal contexts. Notably, performance rankings are inconsistent across $T{=}12$ and $T{=}24$, and no single model dominates across both window lengths and metrics, underscoring that A3 poses a fundamentally harder detection problem than single-vessel anomalies. Excluding Anomaly Transformer, most dedicated anomaly detectors produce near-random AUROC, confirming that standard unsupervised reconstruction objectives are poorly aligned with interaction-driven anomaly cues. These findings motivate the pairwise representation experiment discussed below.

Since a single-trajectory segment cannot fully capture the relational nature of inter-vessel anomalies, we conduct an additional experiment that explicitly encodes pairwise vessel interactions. Following prior work on STGVAD~\cite{kim2026stgvad}, we construct paired inputs using two schemes: (i) \emph{concatenation}, which directly concatenates feature vectors of the own ship and the interacting vessel, and (ii) \emph{difference}, which encodes relative-motion cues via feature-wise differences. For the graph-based input, we use a minimal interaction graph between two vessels at time $t$ and $t{+}1$, evaluated using the same GAT~\cite{velivckovic2017graph}+LSTM architecture from STGVAD. Note that since STGVAD operates under a supervised setting with labeled anomalies, this experiment departs from the anomaly-free training protocol used elsewhere and instead adopts a supervised configuration with a 70/15/15 train/validation/test split using ground-truth anomaly labels; thus absolute numbers are not directly comparable to Table~\ref{tab:baseline}. As shown in Table~\ref{tab:a3_exp}, incorporating inter-vessel information improves detection performance in most settings, with STGVAD Diff achieving the best F1 in the majority of configurations. The gains become more pronounced at higher anomaly ratios, confirming that explicitly modeling relative-motion structure is beneficial for robust A3 detection, while variability at lower ratios (1\%) reflects the inherent difficulty of detecting sparse interaction anomalies.

\paragraph{\textbf{Discussion.}}

\begin{table}[h]
\centering
\caption{AUPRC across anomaly ratios (timestamp-level). Bold and \underline{underline} denote the best and second-best per column.}
\label{tab:ratio_auprc}
\resizebox{\columnwidth}{!}{%
\begin{tabular}{clcccccc}
\toprule
\multirow{2.5}{*}{\textbf{Type}} & \multirow{2.5}{*}{\textbf{Model}} & \multicolumn{3}{c}{$\mathbf{T{=}12}$} & \multicolumn{3}{c}{$\mathbf{T{=}24}$} \\
\cmidrule(lr){3-5} \cmidrule(lr){6-8}
& & \textbf{1\%} & \textbf{3\%} & \textbf{5\%} & \textbf{1\%} & \textbf{3\%} & \textbf{5\%} \\
\midrule
\multirow{7}{*}{A1}
 & IForest             & 0.44 & 1.35 & 2.32 & 0.34 & 1.19 & 2.20 \\
 & OmniAnomaly         & 0.35 & 1.25 & 2.12 & 0.31 & 1.12 & 1.98 \\
 & LSTM-VAE            & 0.39 & 1.37 & 2.33 & 0.33 & 1.23 & 2.16 \\
 & Anomaly Transformer & 0.42 & 1.45 & 2.48 & 0.39 & \underline{1.44} & 2.39 \\
 & LSTM                & 0.48 & \underline{1.66} & \underline{2.80} & \underline{0.40} & \textbf{1.49} & \underline{2.58} \\
 & VanillaTransformer  & \underline{0.49} & \underline{1.66} & \underline{2.80} & \textbf{0.41} & \textbf{1.49} & \textbf{2.60} \\
 & Informer            & \textbf{0.50} & \textbf{1.68} & \textbf{2.83} & \textbf{0.41} & \textbf{1.49} & \textbf{2.60} \\
\midrule
\multirow{7}{*}{A2}
 & IForest             & 1.78 & 6.33 & 10.19 & 1.77 & 7.03 & 13.51 \\
 & OmniAnomaly         & 3.14 & 11.56 & 20.76 & 3.51 & 12.33 & \textbf{25.53} \\
 & LSTM-VAE            & 3.46 & 11.58 & 19.55 & 3.59 & \underline{12.57} & 23.11 \\
 & Anomaly Transformer & 1.24 & 4.18 & 6.94 & 0.83 & 2.55 & 4.29 \\
 & LSTM                & \underline{5.02} & \underline{15.72} & \underline{26.35} & 3.47 & 12.44 & 22.13 \\
 & VanillaTransformer  & \textbf{5.46} & \textbf{17.52} & \textbf{28.84} & \textbf{4.08} & \textbf{13.71} & \underline{23.60} \\
 & Informer            & 4.80 & 15.42 & 25.45 & \underline{3.65} & 12.55 & 22.14 \\
\midrule
\multirow{7}{*}{A3}
 & IForest             & 0.50 & 1.44 & 2.35 & 0.23 & 0.68 & 1.13 \\
 & OmniAnomaly         & 0.61 & 1.69 & 3.00 & 0.39 & 0.91 & 1.46 \\
 & LSTM-VAE            & 0.86 & 2.71 & 4.67 & 0.38 & 1.26 & 1.97 \\
 & Anomaly Transformer & \textbf{1.63} & \textbf{4.76} & \textbf{7.88} & 0.22 & 0.65 & 1.09 \\
 & LSTM                & 0.88 & 2.68 & 4.57 & \underline{0.47} & 1.51 & \textbf{2.47} \\
 & VanillaTransformer  & \underline{0.92} & \underline{2.83} & \underline{4.84} & \textbf{0.49} & \textbf{1.56} & \textbf{2.47} \\
 & Informer            & 0.85 & 2.62 & 4.49 & 0.46 & \underline{1.53} & \underline{2.43} \\
\bottomrule
\end{tabular}%
}
\end{table}

As shown in Table~\ref{tab:ratio_auprc}, AUPRC degrades consistently as the anomaly ratio decreases, yet the degree of degradation and model rankings vary markedly by anomaly type. For A1, all models stay uniformly low, with rankings stable across both window lengths. For A2, forecasting-based models outperform classical ones and drop steeply at lower ratios, but their leadership likewise persists across $T{=}12$ and $T{=}24$. For A3, in contrast, model leadership shifts substantially between $T{=}12$ and $T{=}24$.

Overall, detection performance is primarily governed by the match between an anomaly type's defining cues and a model's inductive bias: position-centric anomalies (A1) remain challenging for most baselines as position-only spikes leave kinematic signals intact, kinematic deviations (A2) are more amenable to representation learning as they produce distinctive SOG/COG signals, and near-miss events (A3) expose a clear limitation of sequence-centric modeling as interaction-driven cues require explicit pairwise reasoning beyond single-trajectory representations. These findings justify anomaly-type-specific evaluation and motivate future work on interaction-aware detection architectures.

\section{Conclusion} \label{sec:conclusion}
This study addresses a central challenge in maritime anomaly detection: defining and evaluating domain-defined anomalies in public AIS datasets. We distinguish single-vessel and inter-vessel anomalies and propose an equation-grounded taxonomy operationalizable under a limited AIS schema. We further employ an LLM as a constrained scoring module to generate context-conditioned plausibility scores, which guide an equation-grounded synthesizer to place and label anomalies. Across diverse settings and baselines, we find performance is largely determined by how well a model's assumptions align with anomaly-specific cues, underscoring the advantage of domain-defined supervision over sparsity-based labeling. Taken together, our results introduce an equation-grounded \emph{score--synthesize--label} pipeline and a standardized evaluation framework, while motivating further research on interaction-aware modeling and broader anomaly taxonomies.
\section{GenAI Disclosure}
A large language model was used in two limited capacities: as a constrained scoring module within the anomaly synthesis framework (Section~\ref{sec:methodology}), with all outputs validated against domain-defined thresholds; and for minor manuscript copyediting. All research design, experimental decisions, and scientific interpretations were made solely by the authors.
\section{Limitations and Ethical Considerations} \label{sec:limitations}
Our framework addresses a restricted subset of formalizable anomaly types and does not cover the full spectrum of real-world maritime anomalies. LLM scorer outputs may reflect prompt design choices and dataset-specific priors, motivating future work on more comprehensive taxonomies and semantically grounded labeling strategies. The framework itself is dataset-agnostic: Algorithm~\ref{alg:eq_grounded_labeling} accepts any AIS data with standard fields regardless of geographic region or temporal resolution. Extension to other vessel types and regions remains for future work. The expert evaluation was conducted with informed consent; no personal data was collected.

\begin{acks}
This work was supported by the Korea Research Institute of Ships and Ocean Engineering (KRISO) through the Endowment Project ``Digital Service Development to Support Marine Digital Transformation,'' funded by the Ministry of Oceans and Fisheries, Republic of Korea (Grant No. 2520001041, PES5950); by the Institute of Information \& Communications Technology Planning \& Evaluation (IITP) under the ``Leading Generative AI Human Resources Development'' grant (IITP-2024-RS-2024-00397085) funded by the Ministry of Science and ICT (MSIT); and by the Korea Health Technology R\&D Project through the Korea Health Industry Development Institute (KHIDI), funded by the Ministry of Health \& Welfare, Republic of Korea (Grant No. RS-2025-02307233).
\end{acks}

\bibliographystyle{ACM-Reference-Format}
\bibliography{ref}

\appendix
\section{Additional Details of Baselines}\label{app:ad}

\begin{itemize}
    \item \textbf{LOF}~\cite{breunig2000lof}: Density-based detector that scores each point by comparing its local reachability density to neighboring points.
    \item \textbf{OCSVM}~\cite{scholkopf2001ocsvm}: Learns a decision boundary enclosing normal data in feature space, treating points outside as anomalies.
    \item \textbf{IForest}~\cite{liu2008iforest}: Isolates anomalies via recursive random partitioning, where anomalous points yield shorter path lengths.
    \item \textbf{LSTM-VAE}~\cite{park2018lstmvae}: LSTM-based VAE that flags anomalies via reconstruction probability or reconstruction error.
    \item \textbf{OmniAnomaly}~\cite{su2019omnianomaly}: Probabilistic RNN with stochastic latent variables that detects anomalies via low reconstruction probability.
    \item \textbf{USAD}~\cite{audibert2020usad}: Dual-autoencoder framework with adversarial training to amplify reconstruction errors on anomalies.
    \item \textbf{TranAD}~\cite{tuli2022tranad}: Transformer with focus scores and adversarial training to leverage broader temporal context for detection.
    \item \textbf{Anomaly Transformer}~\cite{xu2022anomaly}: Detects anomalies via prior--series association discrepancy, trained with a minimax objective for normal--abnormal separability.
    \item \textbf{LSTM}~\cite{hochreiter1997lstm}: Recurrent network with gated memory cells for capturing long-term temporal dependencies.
    \item \textbf{VanillaTransformer}~\cite{vaswani2017transformer}: Multi-head self-attention model for parallelizable long-range dependency modeling.
    \item \textbf{Informer}~\cite{zhou2021informer}: Long-sequence Transformer using ProbSparse attention and distillation to reduce complexity.
    \item \textbf{Autoformer}~\cite{wu2021autoformer}: Auto-Correlation model with progressive series decomposition for trend and seasonality modeling.
    \item \textbf{FEDformer}~\cite{zhou2022fedformer}: Seasonal-trend decomposition Transformer with Fourier and Wavelet frequency-domain attention.
    \item \textbf{Pyraformer}~\cite{liu2022pyraformer}: Pyramidal attention model with inter- and intra-scale connections for linear-complexity modeling.
    \item \textbf{PatchTST}~\cite{nie2023patch}: Patch-tokenized Transformer with channel-independent attention for long-horizon forecasting.
    \item \textbf{DLinear}~\cite{zeng2023dlinear}: Decomposition-based linear model that separately forecasts trend and residual components.
    \item \textbf{STGVAD}~\cite{kim2026stgvad}: Constructs spatio-temporal graphs from AIS trajectories and applies graph-based representation learning for vessel anomaly detection.
\end{itemize}

\section{LLM Prompts}
\label{app:prompt}

For all anomaly types (A1--A3), we structure the LLM interaction with a clear separation between a \emph{system prompt} and a \emph{user query}. The system prompt fixes the LLM's role as an anomaly timestamp scoring module and enforces constraints on output format, score semantics, and anomaly-type-specific structural rules. The user query provides the concrete task instance with externally fixed parameters, without allowing the LLM to redefine scenarios or thresholds.

\subsection{Unified Scoring Prompt}

\begin{prompt}{System Prompt}
You are an "Anomaly Timestamp Scoring Module" for AIS trajectory segments.

Your task is NOT to define anomaly scenarios. The target anomaly type is provided in the user query and is already defined by the fixed rules below.

Your ONLY responsibility:
Given a fixed-length trajectory segment (T timestamps), compute a per-timestamp anomaly suitability score in [0,1], which will later be used by an Equation-Grounded Anomaly Synthesizer to decide WHERE to inject anomalies.

The scores represent RELATIVE PRIORITY, not probabilities.

==================================================
GENERAL SETTING
- Number of timestamps: T (e.g., T=24).
- You must output exactly T scores.
- Scores should be interpretable and consistent with the anomaly-type-specific constraints.
- You MUST respect anomaly-type-specific structural constraints below.
- When a target anomaly count K is provided externally, it is FIXED.
- K MUST NOT be inferred, predicted, or modified.

==================================================
ANOMALY-TYPE-SPECIFIC CONSTRAINTS
[A1: Unexpected AIS Activity]
- Nature: position noise / sensor artifact.
- The anomaly count is fixed to the externally provided K.
- Anomalous timestamps:
    - do NOT need to be consecutive.
    - MAY be scattered.
- Score characteristics:
    - spike-like or irregular patterns are allowed.
    - local sharp increases are acceptable.
    - surrounding timestamps may remain low.
- Physical intuition:
    - anomaly suitability increases where implied motion inferred from positions conflicts with stable SOG/COG.
    - temporal smoothness is NOT required.

[A2: Route Deviation]
- Nature: real but atypical maneuver.
- The anomaly count is fixed to the externally provided K.
- Anomalous timestamps are consecutive in time.
- Score characteristics:
    - scores should form a single contiguous high-score region.
    - smooth rise $\rightarrow$ plateau $\rightarrow$ fall patterns are preferred.
    - isolated spikes are NOT allowed.
- Physical intuition:
    - anomaly suitability increases where SOG and/or COG change rapidly relative to their surrounding context.
    - vessel positions remain physically consistent.

[A3: Close Approach]
- Nature: interaction-induced near-miss between two vessels.
- Interpretation: the segment is a short temporal crop from a longer trajectory, chosen such that a single close-approach event is contained within it.
- The event corresponds to the moment when the two vessels are closest, in the sense of a DCPA-like concept.

Physical interpretation of scores:
- A higher score indicates that the timestamp is more representative of the near-miss interaction apex, i.e., when the inter-vessel separation is closest to a reference distance $\tilde{D}$.
- $\tilde{D}$ should be interpreted as an approximate DCPA surrogate, typically within the range of 0.1--0.3 nautical miles.
- The score does NOT encode distance directly, but reflects how suitable a timestamp is for detecting such a close-approach configuration.

==================================================
SCORING RULES
- Output scores in [0,1].
- Scores indicate how strongly a timestamp should be selected by a downstream anomaly synthesizer.
- The absolute scale matters less than RELATIVE ordering.
- Your task is to assign scores such that a downstream localization rule can naturally select the most suitable timestamps.
- Do NOT explicitly mark, identify, or select the top-K timestamps yourself.
- Do NOT threshold or binarize scores.

==================================================
OUTPUT STRICTNESS
- Return JSON ONLY. Do not output any text outside the JSON object.
- Do not add any extra keys beyond the output schema.
- "anomaly_type" must match the target anomaly type provided in the user query.
- "scores" must have length exactly T.
- If K is provided in the user query, include "K" and set it exactly to the externally provided fixed K.
- (Optional) Round scores to 3 decimal places.

==================================================
OUTPUT FORMAT (JSON ONLY)
{
  "T": <int>,
  "anomaly_type": "<A1|A2|A3>",
  "K": <int or null>,
  "scores": [<float>, ..., <float>]
}

==================================================
IMPORTANT
- Do NOT restate anomaly definitions.
- Do NOT invent missing data.
- If information is insufficient, still produce a plausible score shape consistent with the target anomaly type.
- The downstream system will decide thresholds and perform deterministic anomaly synthesis using the returned scores.

When an input segment is provided, output the JSON score vector.
\end{prompt}

\subsection{User Query Template}

\begin{prompt}{User Query}
[Input Trajectory Segment]

Metadata:
- Vessel Type: {vessel_type}
- Duration: {duration}
- Segment length: T = {T}
- Baseline Statistics: Avg SOG, Std SOG, Avg COG, Std COG

Dynamics (Vectorized for T):
- SOG (knots): {sog_sequence}
- COG (degrees): {cog_sequence}
- Latitude: {lat_sequence}
- Longitude: {lon_sequence}

==================================================
[Target Anomaly Type] 
{anomaly_type}

==================================================
[Fixed Target Count] (Applies to A1/A2 only)
- Target anomaly count K is FIXED to {target_k}.
- This value is externally specified and MUST be respected.

==================================================
[Task]
Generate anomaly suitability scores for Anomaly Type: {anomaly_type}.

Focus on identifying timestamps that are most suitable for the target anomaly type, according to the anomaly-type-specific structural rules defined in the system prompt.

Return JSON only.
\end{prompt}

\section{Anomaly Quality Validation}\label{app:validation}

\paragraph{\textbf{Expert Evaluation.}}
To assess the operational plausibility of the synthesized anomalies, we conducted a domain expert evaluation with five researchers from the Korea Research Institute of Ships and Ocean Engineering (KRISO). Each expert independently assessed 30 synthesized samples per anomaly type (90 samples total, 450 ratings), evaluating each sample on two criteria: (i) operational plausibility (whether the anomaly reflects a realistic maritime event) and (ii) severity (whether the perturbation magnitude is appropriate or too severe). As shown in Table~\ref{tab:expert_eval}, A3 achieves 100.0\% operational plausibility, while A1 and A2 reach 76.7\% and 83.3\%, respectively. A3 shows higher severity ratings (50.0\% ``too severe''), consistent with its DANGEROUS-range DCPA values ($\widetilde{D} \in [0.1, 0.3]$~NM) being at the boundary of operationally critical thresholds.

\begin{table}[h]
\centering
\caption{Expert evaluation results for synthesized anomalies. Plausibility rates are computed over majority-plausible samples; A1 and A2 rates include `Uncertain' responses alongside `No' responses.}
\label{tab:expert_eval}
\resizebox{\columnwidth}{!}{%
\begin{tabular}{lccc}
\toprule
\textbf{Type} & \textbf{Operationally Plausible (\%)} & \textbf{Appropriate (\%)} & \textbf{Too Severe (\%)} \\
\midrule
A1 & 76.7 & 69.4 & 30.6 \\
A2 & 83.3 & 62.5 & 37.5 \\
A3 & 100.0 & 50.0 & 50.0 \\
\bottomrule
\end{tabular}%
}
\end{table}

\paragraph{\textbf{Geographic Validation.}}
Geographic plausibility of A3 is further confirmed: we verified synthesized points against two coastline datasets (Natural Earth 10m, GSHHG), finding that over 98.6\% of A3 points remain at sea ($T{=}12$: 0.18\% on-land; $T{=}24$: 1.38\% on-land), with all flagged cases attributable to coastline polygon approximation rather than inland trajectories. Together, these results support that the synthesized anomalies reflect meaningful real-world abnormal behaviors rather than merely satisfying the synthesis equations.

\clearpage
\end{document}